\documentclass[10pt,twocolumn,letterpaper]{article}

\usepackage[pagenumbers]{cvpr} 
\usepackage{booktabs}
\usepackage{multirow}
\usepackage{colortbl}

\usepackage{graphicx} 
\usepackage{makecell} 
\definecolor{mygray}{gray}{.9}

\usepackage[dvipsnames]{xcolor}

\definecolor{cvprblue}{rgb}{0.21,0.49,0.74}
\usepackage[pagebackref,breaklinks,colorlinks,citecolor=cvprblue]{hyperref}

\def\paperID{5} 
\def\confName{CVPR}
\def\confYear{2024}

\title{Car-1000: A New Large Scale Fine-Grained Visual Categorization Dataset}

\author{Yutao Hu$^{1}$\footnotemark[1], \quad Sen Li$^{2}$\footnotemark[1], \quad Jincheng Yan$^{2}$\footnotemark[1], \quad Wenqi Shao$^{3}$, \quad Xiaoyan Luo$^{2}$\footnotemark[2]\\
$^{1}$Key Laboratory of New Generation Artificial Intelligence Technology and \\ Its Interdisciplinary Applications (Southeast University), Ministry of Education
 \\ $^{2}$ School of Astronautics, Beihang University, Beijing, China  \quad $^{3}$Shanghai AI Laboratory  
}

\begin{document}
\maketitle

\footnotetext[1]{Equal contribution.}
\footnotetext[2]{Corresponding author.}

\begin{abstract}
Fine-grained visual categorization (FGVC) is a challenging but significant task in computer vision, which aims to recognize different sub-categories of birds, cars, airplanes, etc. Among them, recognizing models of different cars has significant application value in autonomous driving, traffic surveillance and scene understanding, which has received considerable attention in the past few years. However, Stanford-Car, the most widely used fine-grained dataset for car recognition, only has 196 different categories and only includes vehicle models produced earlier than 2013. Due to the rapid advancements in the automotive industry during recent years, the appearances of various car models have become increasingly intricate and sophisticated. Consequently, the previous Stanford-Car dataset fails to capture this evolving landscape and cannot satisfy the requirements of automotive industry. To address these challenges, in our paper, we introduce Car-1000, a large-scale dataset designed specifically for fine-grained visual categorization of diverse car models. Car-1000 encompasses vehicles from 166 different automakers, spanning a wide range of 1000 distinct car models. Additionally, we have reproduced several state-of-the-art FGVC methods on the Car-1000 dataset, establishing a new benchmark for research in this field. We hope that our work will offer a fresh perspective for future FGVC researchers. Our dataset is available at \url{https://github.com/toggle1995/Car-1000}.

\end{abstract}    
\section{Introduction}
\label{sec:intro}

Fine-grained visual categorization \cite{lin2015bilinear, sun2018multi, zheng2019looking, he2022transfg, hu2021alignment} aims to recognize objects from different subordinate categories, \emph{e.g.}, species of birds, models of cars, or different brands of planes. Among them, accurately recognizing different models of cars has received extensive attention in recent years due to their large value in various applications such as autonomous driving and traffic surveillance \cite{xiang2019global, hu2023planning}. However, owing to the extreme similarity among different models of cars, the easily extracted global features, such as structure, color and shape contribute minimally to precise identification \cite{chen2019destruction, du2020fine}. To achieve accurate recognition, more efforts should be made to extract detailed local information, which calls for meticulous designed methods \cite{chen2019destruction, du2020fine}. 

To explore a robust and practical algorithm towards fine-grained car classification, a comprehensive dataset is important. Unfortunately, Stanford Car \cite{krause20133d}, the most popular dataset in this field, fails to meet this basic requirement. On the one hand, Stanford Car only contains 196 categories, which limits the diversity of the dataset. On the other hand, it only includes the cars released before 2013, which fails to capture the recent trends in the automotive industry. As for other datasets toward fine-grained car classification \cite{dong2015vehicle, yang2015large, tafazzoli2017large}, considering their limited scales, categories and attribute information, none can serve as a comprehensive benchmark. Hence, to support the in-depth research in the fine-grained car classification, there is an urgent need for a new dataset that exhibits stronger representativeness and diversity.

To address the aforementioned problems, in this paper, we introduce Car-1000, a novel and large-scale comprehensive dataset for fine-grained car classification. First, to ensure the dataset aligns with the latest developments in the automotive industry, we extract the popularity and user comments of different car models from one of the world's largest automotive forums, ``DongCheDi'' \cite{dongchedi2024}, and select 1000 high-attention categories to be included in our dataset. Then, we acquire images of each category from the internet through web scraping scripts and employ professional annotators with considerable knowledge about cars to meticulously filter the images. The whole process cost us more than 4000 dollars. Consequently, we present a new, large-scale fine-grained car dataset containing 1000 categories and 140267 images. The whole dataset has been released to the community, thus facilitating future research endeavors in this field.

Generally speaking, the proposed Car-1000 dataset is distinguished by three advantages:

\begin{enumerate}	
\item Our Car-1000 dataset is a large-scale comprehensive dataset, comprising 140267 images across 1000 different models from 166 automakers. Considering both the total number of classes, the volume of images, and the diversity of the included automakers, our Car-1000 emerges as the largest and most comprehensive dataset towards fine-grained car classification.

\item  Besides the 1000 categories, the Car-1000 incorporates a three-tier hierarchical label system that captures the attributes and types of different vehicles. At the first level,we divide all entries into 7 primary-categories, \emph{e.g.,} sedan, truck, sports car, bus, van, MPV, and SUV. Subsequently, within each primary-category, we continually classify the vehicles into secondary categories based on their sizes, thus establishing a tree-like structure with 7 primary and 21 secondary categories. The specific structure of the label system is listed in Table~\ref{table:tree}.

\item Car-1000 dataset aligns with recent trends in the automotive industry. As shown in Fig.~\ref{fig:time}, our collection encompasses a broader coverage of vehicles released from the 1960s to the 2020s, offering a more comprehensive overview than the prior datasets. This broader temporal coverage ensures our dataset more effectively supports current application requirements in the automotive field.

\end{enumerate}

Furthermore, to establish a reference benchmark for future research, we reproduce 16 different classification networks on Car-1000 and evaluate their performance. The 16 selected networks include 7 general classification models and 9 fine-grained classification ones. The experimental results reveal that our Car-1000 is extremely challenging, with no network delivering accuracy higher than 90\%. A more detailed analysis is provided in the Sec.~\ref{sec:exp}.

\section{Car-1000 Dataset}
\label{sec:dataset}

\subsection{Data Collection}

To develop the comprehensive Car-1000 dataset, it is essential to encompass a diverse range of vehicles from various manufacturers. Meanwhile, to meet the application demand, our Car-1000 dataset should be consistent with the market interest. For this purpose, we collect information of different car models from DongCheDi \cite{dongchedi2024}, one of the largest automotive websites. DongCheDi provides detailed insights on various car models, including comments and user evaluations, which reflect the popularity of each model. We hypothesize that the greater the popularity, the more significant the car model is for applications. Therefore, we select 1000 car models based on their popularity on DongCheDi to obtain the category list of Car-1000.

Based on the obtained category list, we utilize web scraping techniques to collect images from the internet. We aim to collect 500 images for each car model, resulting in a total of 500,000 raw images. Subsequently, we apply the MD5 hash algorithm \cite{rivest1992md5} to eliminate duplicate images within each class. By doing so, we obtain 394871 unique images without duplication, which will be delivered to professional annotators for further refinement.

Specifically, we engage three annotators, each with extensive expertise in the automotive domain, to manually filter the images collected through web scraping. Specifically, we evenly divide the dataset into three parts, denoted as Part-A, Part-B, and Part-C. Each annotator is assigned two different parts for review. During the annotation process, if two annotators agree on the opinion of an image, whether to retain or exclude it, the decision will be made based on their consensus. Conversely, if the opinions of two annotators are divergent, the third annotator is consulted to make the final decision. This process is significant for developing a high-quality dataset, which costs us more than 4000 dollars.

\begin{figure}[tbp]
\centering 
\includegraphics[width=1\linewidth]{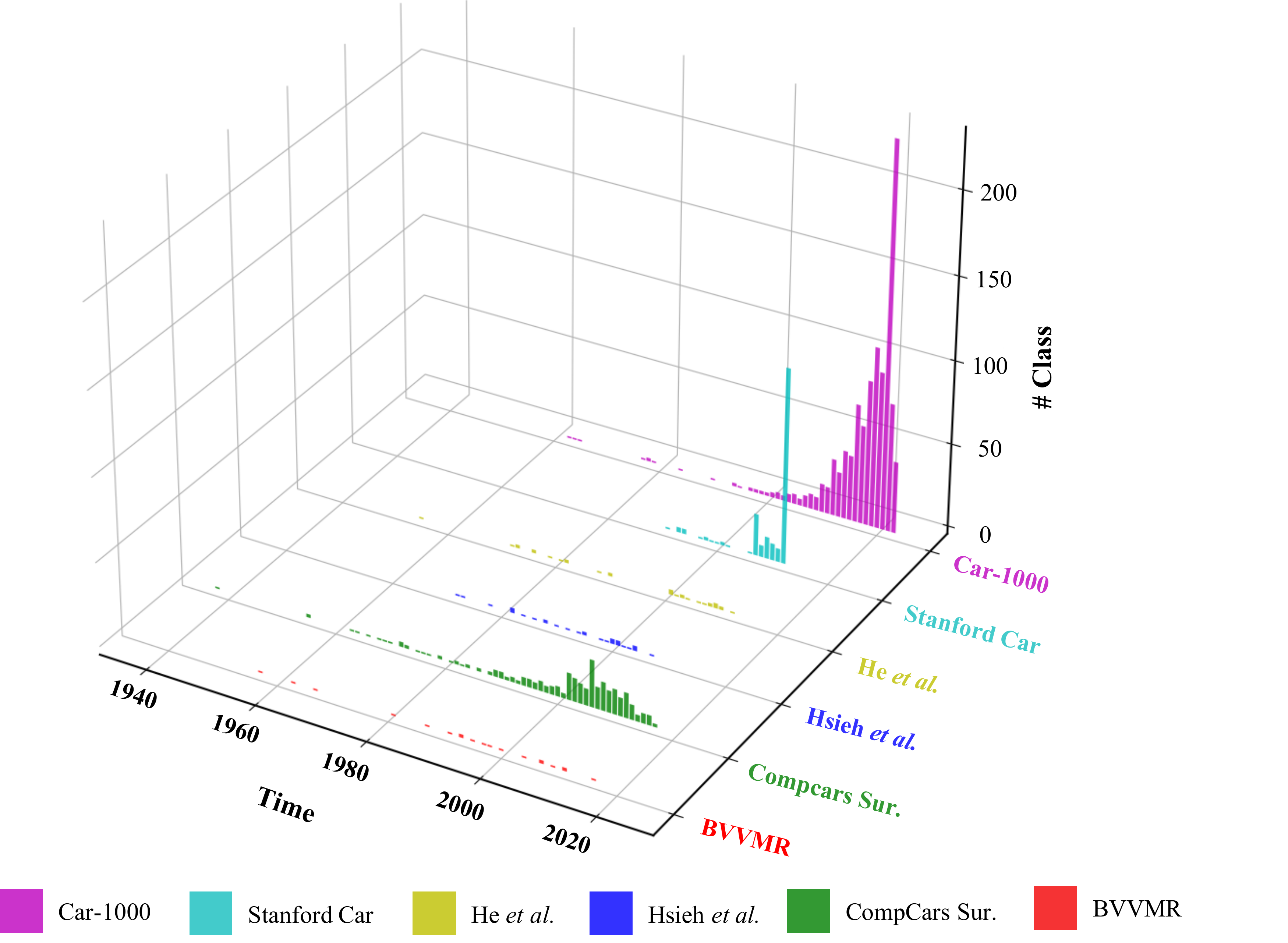} 
\caption{We depict the temporal coverage of car models involved in Stanford Car \cite{krause20133d}, He \emph{et al,} \cite{he2015recognition}, Hsieh \emph{et al,} \cite{hsieh2014symmetrical}, CompCars Sur.\cite{yang2015large}, BVVMR \cite{biglari2017part} and our Car-1000. The higher the column is, the more models released in that years are included in corresponding dataset. It is obvious that Car-100 has a wide temporal coverage and contains the new models released in recent years.}
\label{fig:time} 
\end{figure}

Finally, to protect the privacy of car owners, we obscure the license plates of the car that appeared in the image. The region of license plates will be masked with an RGB value of [170, 180, 190].

\begin{figure*}[tbp]
\centering 
\includegraphics[width=1\linewidth]{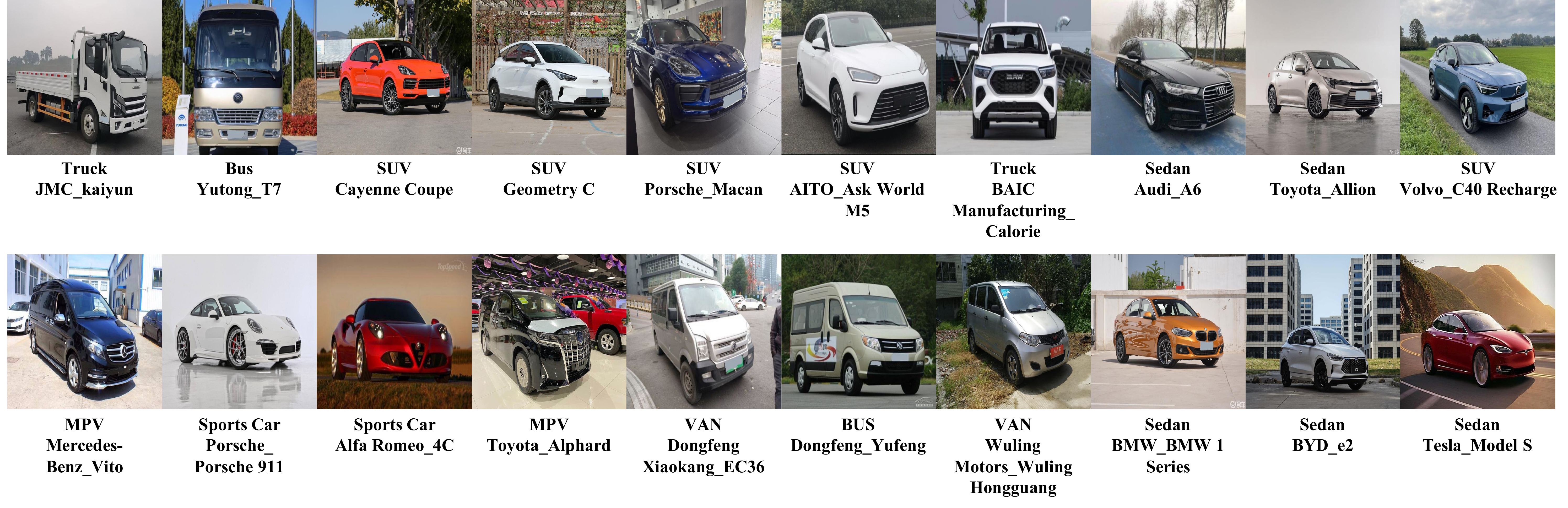} 
\vspace{-8mm}
\caption{We present some selected samples from our Car-1000 dataset, with the primary label and the name of the corresponding model provided below each image. Notably, the names before and after ``\_'' indicate the automakers and the specific model, respectively.}
\label{fig:dataset} 
\end{figure*}

\subsection{Dataset Characteristics}

\textbf{Relative Large-scale dataset.} Our Car-1000 encompasses 140267 images, spanning 1000 categories from 166 automakers. Considering the number of automakers, images, classes and the average images per class, our Car-1000 stands out as the most comprehensive dataset, which surpasses the previous ones such as Stanford Cars \cite{krause20133d}, CompCars \cite{yang2015large}, BVMMR \cite{biglari2017part}.

\begin{table}[tbp]
\centering
\caption{The structure of our Car-1000 Dataset. The ``\# Class'' indicates the sub-class number of the corresponding super-class.}
\vspace{-2mm}
\begin{tabular}{|c|c|c|c|}
\hline
\multicolumn{3}{|c|}{Super-Class} & \# Classes \\ 
\hline
\multirow{21}{*}{\rotatebox[origin=c]{90}{Car-1000 Dataset}} & \multirow{6}{*}{Sedan} & Large Sedan & 22 \\
\cline{3-4}
&                          & Compact Sedan & 96 \\
\cline{3-4}
&                          & Micro Sedan & 35 \\
\cline{3-4}
&                          & Small Sedan & 21 \\
\cline{3-4}
&                          & Mid-size Sedan & 75 \\
\cline{3-4}
&                          & Upper Mid-size Sedan & 49 \\
\cline{2-4}
& \multirow{3}{*}{Truck} & Pickup Truck & 53 \\
\cline{3-4}
&                          & Light Truck & 7 \\
\cline{3-4}
&                          & Micro Truck & 41 \\
\cline{2-4}
& \multirow{1}{*}{Sports Car} & Sports Car & 44 \\
\cline{2-4}
& \multirow{1}{*}{Bus} & Bus & 32 \\
\cline{2-4}
& \multirow{1}{*}{Van} & Van & 34 \\
\cline{2-4}
& \multirow{4}{*}{MPV} & Compact MPV & 22 \\
\cline{3-4}
&                          & Small MPV & 1 \\
\cline{3-4}
&                          & Mid-size MPV & 7 \\
\cline{3-4}
&                          & Upper Mid-size  MPV & 39 \\
\cline{2-4}
& \multirow{5}{*}{SUV} & Large SUV & 18 \\
\cline{3-4}
&                          & Compact SUV & 154 \\
\cline{3-4}
&                          & Small SUV & 61 \\
\cline{3-4}
&                          & Mid-size SUV & 109 \\
\cline{3-4}
&                          & Upper Mid-size  SUV & 80 \\
\hline
\end{tabular}
\label{table:tree}
\end{table}

\textbf{Hierarchical Labeling System.} To enrich the attribute information for each category in our Car-1000 dataset, we have developed a three-tier hierarchical label-tree. As listed in Table~\ref{table:tree}, we initially categorize the vehicles into 7 different primary categories, \emph{e.g.,} sedan, truck, sports car, bus, van, Multi-Purpose Vehicle (MPV) and Sport Utility Vehicle (SUV). Then, within sedan, truck, MPV and SUV, we further build secondary super-classes based on the size of the vehicles, leading to 21 distinct secondary super-classes. In this way, we provide more attribute information for each car model, establishing a more comprehensive and informative dataset than previous datasets. The number of different car models included in each super-class is shown in Table~\ref{table:tree}.

\textbf{Broader Temporal Coverage.} Compared to previous datasets, our dataset has a relatively broader temporal coverage. Fig.~\ref{fig:time} displays the year-wise distribution of car models included in Stanford Car \cite{krause20133d}, He \emph{et al.} \cite{he2015recognition}, Hsieh \emph{et al.} \cite{hsieh2014symmetrical}, CompCars Sur. \cite{yang2015large}, BVVMR \cite{biglari2017part} and our Car-1000. Each column represents the quantity of different models released in a specific year in the corresponding dataset. We can find that our Car-1000 has the broadest temporal coverage. Meanwhile, it is obvious that Car-1000 is the most up-to-date dataset, containing 640 car models released in recent 5 years, especially 450 ones after 2020. This characteristic allows our dataset to more effectively capture the latest development trends in the automotive industry.

\section{Experiments}
\label{sec:exp}

\begin{table*}[tbp]
  \centering
  \small
  \caption{The performance comparison of different classification networks in terms of OA (\%), AA (\%) and Kappa (\%) on val set and test set, respectively. The whole table is split into two parts over the rows. The first part reports the accuracy of general-purpose networks while the second parts presents the performance of fine-grained specific methods. In each part, the best and second-best performance are marked in red and blue, respectively.}
  \vspace{-2mm}
    \begin{tabular}{l|c|c|c|c|c|c|c|c}
    \toprule[1pt]
    \multirow{2}{*}{Method} & \multirow{2}{*}{\makecell[c]{Backbone}} & \multirow{2}{*}{\makecell[c]{\# Params}} & \multicolumn{3}{c|}{Val Set} & \multicolumn{3}{c}{Test Set} \\
\cline{4-9} &  &  & OA   & AA  & Kappa  & OA   & AA  & Kappa  \\
    \hline
    VGG-19 \cite{szegedy2015going} & VGG-19 &  20.56M & 85.16 & 82.97 & 85.14 & 85.55 & 83.36 & 85.53 \\
    ResNet-50 \cite{he2016deep} & ResNet-50  & 27.66M  & 84.38 & 82.11 & 84.36 & 84.57 & 82.40 & 84.55 \\
    SENet-50 \cite{hu2018squeeze} & SENet-50  & 30.14M  & 84.77  & 82.60  & 84.76 & 84.71  & 82.56 & 84.70 \\
    DenseNet-169 \cite{huang2017densely} & DenseNet-169  & 14.15M  & \textcolor{red}{85.90} & \textcolor{red}{83.87} & \textcolor{red}{85.89}  & \textcolor{red}{86.09}  & \textcolor{red}{84.00}  & \textcolor{red}{86.08} \\
    ResNest-50 \cite{zhang2022resnest} & ResNest-50  & 27.48M  & \textcolor{blue}{85.48}  & \textcolor{blue}{83.67} & \textcolor{blue}{85.46}  & \textcolor{blue}{85.77}  & \textcolor{blue}{83.78} & \textcolor{blue}{85.76} \\
    ViT\_B\_16 \cite{dosovitskiy2020image} & Vit\_Base\_16  & 86.57M  & 78.84  & 77.02  & 78.81  & 79.27 & 77.22 & 79.24 \\    
    Swin-Base \cite{liu2021swin} & Swin-Base  & 87.77M  & 83.20  & 81.22 & 83.18  & 82.97  & 81.11 & 82.95 \\
    \hline
    \hline
    Bilinear CNN\cite{lin2015bilinear} & VGG-19  & 282.18M  & 86.71  & 83.93  & 86.17 & 86.74  & 84.17  & 86.31 \\
    S3N \cite{ding2019selective} & ResNet-50  & 113.40M  & 87.51  & 85.72  & 87.49 & 87.73  & 85.80  & 87.72 \\
    DCL \cite{chen2019destruction} & ResNet-50  & 29.70M  & 86.00  & 84.41  & 85.98  & 86.60  & 84.87  & 86.58 \\
    MC-Loss \cite{chang2020devil} & ResNet-50  &  23.50M & 83.20  & 81.04  & 83.18 & 83.63  & 81.18  & 83.61 \\
    PMG \cite{du2020fine} & ResNet-50  & 46.80M  & \textcolor{blue}{88.51}  & \textcolor{blue}{86.42}  & \textcolor{blue}{88.50}  & \textcolor{blue}{88.60}  & \textcolor{blue}{86.39}  & \textcolor{blue}{88.59} \\
    HSD \cite{hu2021hierarchical} & ResNet-50  & 29.40M  & 86.83  & 84.91  & 86.82  & 86.76  & 84.81  & 86.75 \\
    CAL \cite{rao2021counterfactual} & ResNet-101  & 108.10M  & \textcolor{red}{89.07} & \textcolor{red}{87.54} & \textcolor{red}{89.05} & \textcolor{red}{89.45} & \textcolor{red}{87.60}  & \textcolor{red}{89.44} \\
    TransFG \cite{he2022transfg} &  Vit\_Base\_16 & 87.50M & 83.37  & 81.76  & 83.35  & 83.60 & 81.70 & 83.58 \\
    FGVC-PIM \cite{chou2022novel} & Swin-Tiny  & 263.10M  & 87.78  & 86.36  & 87.77  & 88.17 & 86.36  & 88.16 \\
    \bottomrule[1pt]
    \end{tabular}%
  \label{tab:acccom1}%
\vspace{-3mm}
\end{table*}%

\subsection{Implementation Details}
We define a standard split for training, validation (val) and test set within our Car-1000 dataset. Specifically, we adopt a random selection process, extracting 60\%, 20\%, and 20\% of samples from each category to formulate the training, val, and test sets, respectively.  

To establish a robust benchmark, we evaluate 16 different classification networks on our Car-1000 dataset, incorporating both general-purpose and fine-grained-specific networks. Specifically, for general-purpose classification networks, we include five CNN-based networks, \emph{e.g.,} VGGNet \cite{szegedy2015going}, ResNet \cite{he2016deep}, SENet \cite{hu2018squeeze}, DenseNet \cite{huang2017densely} and ResNest \cite{zhang2022resnest}. Meanwhile, two transform-based networks, ViT \cite{dosovitskiy2020image} and Swin-Transformer \cite{liu2021swin}, are also included. In the domain of fine-grained classification, we select 9 distinct networks for evaluation, \emph{e.g.,} Bilinear CNN \cite{lin2015bilinear}, S3N \cite{ding2019selective}, DCL \cite{chen2019destruction}, MC-Loss \cite{chang2020devil}, PMG \cite{du2020fine}, HSD \cite{hu2021hierarchical}, CAL \cite{rao2021counterfactual}, TransFG \cite{he2022transfg} and FGVC-PIM \cite{chou2022novel}.

To better evaluate the classification performance of selected baseline methods within the experiment, we adopt three different metrics, overall accuracy (OA), average accuracy (AA) and kappa coefficient (Kappa). Specifically, the OA (\%) represents the proportion of correctly classified samples to the total number of testing samples. The AA (\%) calculates the mean accuracy across all categories. The kappa coefficient is generally used to measure the agreement between the predicted and actual values. Larger values of OA, AA and Kappa suggest the better performance.

\subsection{Experimental Results}
We reproduce 16 different classification networks and report their accuracy on Table~\ref{tab:acccom1}. Among the \emph{general-purpose networks}, DenseNet-169 achieves the best performance across all metrics on both the validation and test sets with the  fewest parameters. ResNest-50 obtains the second-best performance, surpassing ResNet-50 and SENet-50 with fewer parameters. Additionally, VGG-19 also delivers satisfactory performance with fewer parameters. Based on these experiments, we can find the large model not always lead to better accuracy. Meanwhile, transformer-based network, which plays a dominant role in nowadays computer vision community, does not exhibit superior performance than classical CNN-based networks.

Among the \emph{fine-grained specific networks}, CAL \cite{rao2021counterfactual} achieves the best performance, with PMG \cite{du2020fine} closely following, showcasing superior results across three metrics on both val and test sets. Moreover, among the methods adopting ResNet-50 as the backbone network, PMG exhibits the highest accuracy, highlighting the benefits of its progressive learning strategy on jigsaw patches. Through each jigsaw patch, PMG could better localize the significant local region and extract discriminative features. Additionally, HSD \cite{hu2021hierarchical} achieves relatively superior performance with a smaller parameters size, which reflects the remarkable efficacy of self-distillation in enhancing feature learning.

\section{Conclusion}
\label{sec:conclusion}
In this paper, we introduce Car-1000, a new large-scale comprehensive fine-grained car classification dataset. Compared to the existing datasets, our Car-1000 stands out for the following advantages. First, Car-1000 is one of the largest datasets dedicated to car classification with 140267 images, covering 1000 different models. It is worthwhile to mention that these 1000 different models come from 166 different automakers, endowing Car-1000 with a high level of diversity. Second, Car-1000 exhibits extensive temporal coverage, including the models from 1960s to the latest 2020s. By doing so, Car-1000 dataset is the most up-to-date one in this field, which captures the new trend of car industry. Third, Car-1000 incorporates a three-tier hierarchical labeling system with 7 primary-categories, 21 secondary-categories and 1000 sub-classes, which provides the attribute and size information of each involved model. Additionally, we benchmark several state-of-the-art classification networks on Car-1000, which serves as the baseline results for future researchers. We hope our work will provide a new perspective for future research of FGVC.

{
    \small
    \bibliographystyle{ieeenat_fullname}
    \bibliography{main}
}


\end{document}